 \documentclass[pmlr,twocolumn,10pt]{jmlr} 

\usepackage{booktabs}
\usepackage{siunitx}

\usepackage{pgfplots}
\usepackage{caption} 
\usepackage{tikz} 
\usepackage{enumitem}

\theorembodyfont{\upshape}
\theoremheaderfont{\scshape}
\theorempostheader{:}
\theoremsep{\newline}

\jmlrvolume{LEAVE UNSET}
\jmlryear{2023}
\jmlrsubmitted{LEAVE UNSET}

\jmlrpublished{LEAVE UNSET}

\jmlrworkshop{Machine Learning for Health (ML4H) 2023} 

\title[Outlier detection in RCTs]{Do Ensembling and Meta-Learning Improve Outlier Detection in Randomized Controlled Trials?}

\author{%
\Name{Walter Nelson}\textsuperscript{1,2} \Email{nelsonwa@hhsc.ca}\\
\Name{Jonathan Ranisau}\textsuperscript{1} \Email{ranisau@hhsc.ca}\\
\Name{Jeremy Petch}\textsuperscript{1,3,4,5} \Email{petchj@hhsc.ca}\\
\addr \textsuperscript{1}Centre for Data Science and Digital Health, Hamilton Health Sciences, Hamilton, Canada\\
\addr \textsuperscript{2}Department of Statistical Sciences, University of Toronto, Toronto, Canada\\
\addr \textsuperscript{3}Institute for Health Policy, Management and Evaluation, University of Toronto, Toronto, Canada \\
\addr \textsuperscript{4}Department of Medicine, McMaster University, Hamilton, Canada \\
\addr \textsuperscript{5}Population Health Research Institute, Hamilton, Canada
}

\begin{document}

\maketitle

\begin{abstract}
Modern multi-centre randomized controlled trials (MCRCTs) collect massive amounts of tabular data, and are monitored intensively for irregularities by humans. We began by empirically evaluating 6 modern machine learning-based outlier detection algorithms on the task of identifying irregular data in 838 datasets from 7 real-world MCRCTs with a total of 77,001 patients from over 44 countries. Our results reinforce key findings from prior work in the outlier detection literature on data from other domains. Existing algorithms often succeed at identifying irregularities without any supervision, with at least one algorithm exhibiting positive performance 70.6\% of the time. However, performance across datasets varies substantially with no single algorithm performing consistently well, motivating new techniques for unsupervised model selection or other means of aggregating potentially discordant predictions from multiple candidate models. We propose the \textbf{Me}ta-learned \textbf{P}robabilistic \textbf{E}nsemble (MePE), a simple algorithm for aggregating the predictions of multiple unsupervised models, and show that it performs favourably compared to recent meta-learning approaches for outlier detection model selection. While meta-learning shows promise, small ensembles outperform all forms of meta-learning on average, a negative result that may guide the application of current outlier detection approaches in healthcare and other real-world domains.
\end{abstract}

\section{Introduction}
\label{sec:intro}

\begin{table*}
\centering
\begin{tabular}{lrrrr|rrr}
\toprule
trial & participants & centres & countries & datasets & instances & features & irregular \\
& (n) & (n) & (n) & (n) & (n, avg.) & (n, avg.) & (\%, avg.) \\
\midrule
compass & 27395 & 602 & 33 & 213 & 8188.9 & 60.7 & 16.0 \\
hipattack & 2970 & 69 & 17 & 22 & 676.5 & 61.1 & 32.1 \\
hope3 & 12705 & 228 & 21 & 56 & 4474.2 & 48.6 & 15.4 \\
manage & 1754 & 84 & 19 & 190 & 345.4 & 65.7 & 24.2 \\
poise & 8351 & 190 & 23 & 87 & 3462.2 & 25.3 & 5.4 \\
rely & 18113 & 951 & 44 & 146 & 14150.0 & 40.0 & 17.3 \\
tips3 & 5713 & 86 & 9 & 124 & 3682.0 & 67.0 & 9.2 \\
\midrule
\textbf{overall} & 77,001 & - & - & 838 & 5846.0 & 54.7 & 16.4 \\
\bottomrule
\end{tabular}

\caption{Characteristics of each trial (left) and the average characteristics of their respective datasets (right).}
\label{table:descriptive}
\end{table*}

The practice of evidence-based medicine hinges on the quality of data collected \citep{rcttrust}.
The large, global, multi-centre randomized controlled trial (MCRCT) is considered the gold standard tool for
treatment evaluation, in part because of the rigorous data quality controls put in place \citep{buyse2020}. One such control is
centralized statistical monitoring (CSM), whereby statisticians employed by the trial runner monitor data from
all incoming centres for irregularities using statistical tests and other manual checks \citep{kirkwood2013}. Such processes are
time-intensive and often guided by prior knowledge of previous irregularities, raising questions about their
ability to adapt to new types of irregularities. Machine learning 
represents an interesting opportunity for more flexible irregularity detection. Prior work has shown that
machine learning can successfully identify so-called ``collective anomalies" (i.e., irregular centres) in MCRCTs when traditional statistical approaches cannot \citep{petch2022}.

In this work, we frame the problem of irregularity detection in MCRCTs
as the more classical task of instance-level outlier detection (i.e., identifying anomalous
individual records). Under this formulation, the task is completely unsupervised: the only available
input is the full dataset, and the aim of outlier detection is to identify which instances in the
dataset are irregular (or anomalous). Recent work on instance-level outlier detection for tabular
datasets includes ADBench \citep{han2022adbench}, a comprehensive suite of benchmarks for outlier detection
and other forms of the anomaly detection problem (which incorporate varied levels of supervision
and outlier contamination in the input dataset), along with a re-usable suite of benchmark tabular datasets.
A key finding from ADBench is that performances of well-known outlier detection algorithms vary
significantly across datasets according to the algorithm's inductive biases, the suitability of which to a particular dataset is often not known \textit{a priori}. As a result, the authors highlight the importance of developing new techniques
for unsupervised outlier model selection. Such recent tools as MetaOD \citep{zhao2020automating} and ELECT \citep{zhao2022unsupervised}
fall into this category, using dataset-level hand-crafted meta-features to predict which outlier detection models will perform the
best, allowing for unsupervised model selection when encountering a new dataset.

\noindent \textbf{Contributions.} As no benchmarks of outlier detection models in healthcare are available, we begin with an empirical evaluation of well-known outlier detection algorithms on our large dataset-of-datasets from MCRCTs. The goal is to quantify the performance variability across algorithms and datasets from recent large MCRCTs. Finding that performance varies substantially across algorithms and datasets, we propose the meta-learned probabilistic ensemble (MePE), which combines meta-learning and ensembling and outperforms earlier meta-learning approaches in a head-to-head comparison, but still falls short of ensembling. We conclude with general recommendations for the application of outlier detection models in the context of MCRCTs and open research questions.

\noindent \textbf{Code and data availability.} Our code is made available on GitHub at \url{https://github.com/hamilton-health-sciences/ml4h-traq}. The data from MCRCTs is managed by the investigators who led each respective trial.

\section{Methods}
\label{sec:methods}

\subsection{Data acquisition and preprocessing}

Our study evaluates algorithms on data from several completed MCRCTs: Perioperative Ischemic Evaluation Study (POISE, \cite{poise}), Randomized Evaluation of Long Term Anticoagulant Therapy (RE-LY, \cite{rely}), Heart Outcomes Prevention Evaluation-3 (HOPE-3, \cite{hope3}), The International Polycap Study 3 (TIPS-3, \cite{tips3}), Rivaroxaban for the Prevention of Major Cardiovascular Events in Coronary or Peripheral Artery Disease (COMPASS, \cite{compass}), Hip Fracture Accelerated Surgical Treatment and Care Track (HIP ATTACK, \cite{hipattack}), and Management of Myocardial Injury After Noncardiac Surgery (MANAGE, \cite{manage}).

We worked with statisticians
familiar with the conduct of the trials to develop a procedure to ascertain irregularity labels. For each trial,
we obtained several snapshots over the course of the trial from before trial completion.
Each snapshot contains multiple datasets, each representing a separate case report
form (CRF): for example, one dataset might contain data collected from a CRF completed at the time of patient
enrolment in the trial, while other datasets contain data collected from CRFs for adverse events which can occur
at any time over the course of the trial. We identify irregularities in the preliminary snapshots by comparing
the data with the data in the final, locked trial database. When differences are present, we assume that they
are the result of human-driven quality assurance procedures undertaken between the preliminary snapshot and trial completion. For each instance
in each dataset from each preliminary snapshot with at least one field which differs, we consider the instance to be irregular.

In all experiments, missing values are imputed with the most frequent value in that column. No other preprocessing is
applied to the input data. In all experiments, MetaOD meta-features are clipped to $[0, 1]$ to match the pre-trained model from
the original paper \citep{zhao2020automating}.

\begin{figure}
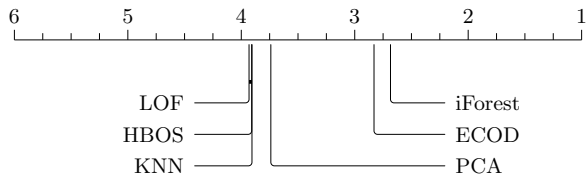

  \includeteximage[width=8cm]{images/exp1cd.tex}
  \caption{Critical differences between the AUROCs of all algorithms across all datasets in our study.}

  \label{fig:exp1cd}
\end{figure}

\subsection{Benchmarking: outlier detection}

We evaluate the ability of existing outlier detection algorithms to identify irregular instances in a fully unsupervised
fashion. Each dataset from each snapshot is provided as input to each algorithm independently, and the areas under the receiver
operating characteristic (AUROC) and precision-recall (AUPR) curves are computed. The algorithms are used with the default hyperparameters given in PyOD \citep{pyod}: Isolation Forest (iForest; \cite{iforest}), Empirical Cumulative Distribution Function-based Outlier Detection (ECOD; \cite{ecod}), $K$-Nearest Neighbours-based Outlier Detection (KNN; \cite{knnod}), Local Outlier Factor (LOF; \cite{lof}), Principal Components Analysis-based Outlier Detection (PCA; \cite{pcaod}), and Histogram-Based Outlier Scores (HBOS; \cite{hbos}).

We report the fraction of datasets where at least one outlier detection algorithm achieves an AUROC
better than a random classifier ($0.5 <$ lower bound of 95\% CI). In outlier detection evaluation, AUROC has a convenient interpretation as the probability
that an irregular instance is predicted as more anomalous than a regular instance \citep{Hand2009}. On each dataset from all trials and snapshots,
we assign each algorithm a rank according to AUROC (with 1 being the best) and report the mean ranks across all datasets graphically on a critical difference diagram \citep{cd}, which connects statistically indistinguishable methods with a thick dark line. We also provide per-trial and per-snapshot
average performances of the overall best algorithm (Appendix \ref{app:eval}). 

\subsection{Model selection}

The ability to provide a single, actionable set of predictions is a critical task
in real-world application of outlier detection models. We conducted an experiment to evaluate techniques for zero-shot model selection \citep{zsautoml} (the task of
choosing the best outlier detection model for a particular dataset, without access to any labels) and ensembling
(aggregating the potentially discordant outputs of several outlier detection models on a particular dataset, again
without access to any labels).

We included five approaches in our evaluation:
\begin{itemize}[noitemsep,topsep=0pt]
    \item MetaOD-P: A MetaOD model pre-trained on a number of publicly available outlier detection
datasets.
    \item MetaOD-R: A MetaOD model re-trained on MCRCT data and algorithm performances (AUROCs) from our first experiment and evaluated in a leave-one-trial-out fashion.
    \item Ensemble-N: A naive ensemble of the decision scores of each outlier detection model.
    \item Ensemble-P: A ``probabilistic" ensemble, which scales the decision scores of each outlier detection model to $[0, 1]$ prior to averaging them \citep{pyod}.
    \item MePE: A \textbf{me}ta-learned \textbf{p}robabilistic \textbf{e}nsemble, which uses MetaOD-R to predict the top $k$ best models and probabilistically ensembles them.
    \item iForest: The overall best model from our first benchmark experiment.
\end{itemize}

MetaOD-P selects a single model from among a large set of models in its meta-training set \citep{zhao2020automating}. MetaOD-R
selects a single model from among the 6 outlier detection models benchmarked in our first experiment. Ensemble-N and Ensemble-P
ensemble the predictions of all 6 outlier detection models benchmarked in our first experiment. MePE ensembles the MetaOD-R-predicted-top-$k$ (out of 6) outlier detection models benchmarked in our first experiment. MePE with $k = 1$ would
thus be equivalent to MetaOD-R; we do not tune the parameter and take $k = 3$ for all experiments. All base models use the PyOD implementation \citep{pyod}. Further methodological details are available in Appendix \ref{app:methods}.

\begin{figure}
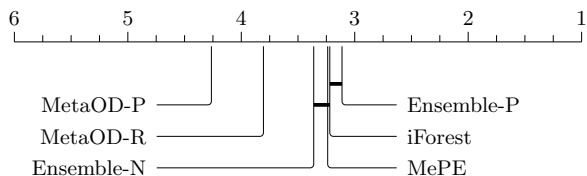

  \includeteximage[width=8cm]{images/exp2cd.tex}
  \caption{Critical differences between the AUROCs of model selection and ensembling techniques across all datasets in our study.}

  \label{fig:exp2cd}
\end{figure}

\section{Results}

Descriptive statistics illustrate the diversity in dataset characteristics across different trials according to the number of instances (rows) and features (columns) in each tabular dataset, along with the proportion of instances that are irregular (Table \ref{table:descriptive}, Appendix \ref{app:data}).

Comparison of outlier detection algorithms reveals that much like in an earlier domain-agnostic benchmark \citep{han2022adbench}, iForest is
the overall best, and a sound choice if a single algorithm must be chosen (Figure \ref{fig:exp1cd}). iForest provides positive performance ($0.5 <$ lower bound of 95\% CI of AUROC) for 60.3\% (505/838) of datasets considered in our study. When it doesn't, another algorithm does 24.9\% of the time. Even though iForest is \textit{a priori} the best option, it is the best performer only 21.0\% of the time; for 79.0\% (662/838) of datasets, the practitioner would have been better off choosing one of the other 5 algorithms considered here. Even the \textit{a priori} worst option, LOF, is the best performer on 11.6\% (97/838) datasets.

To investigate whether it is possible for a practitioner make a more informed choice without access to labels, we
compare recent meta-learning approaches to this \textit{a priori} choice of iForest and various forms of ensembling (Figure \ref{fig:exp2cd} and Table \ref{table:results}). The best performers are probabilistic ensembling and iForest, which are statistically indistinguishable. Although re-training MetaOD on data from MCRCTs boosts performance (MetaOD-R) relative to the pretrained version (MetaOD-P), and combining it with ensembling boosts performance further (our proposed MePE), none beat iForest or a probabilistic ensemble of the 6 models from our initial benchmark, which have the added benefit of simplicity.

\section{Discussion}

To our knowledge, this study is the largest-scale benchmark of tabular outlier detection algorithms (and the first to examine the viability of both ensembling and meta-learning) in real-world datasets from any domain. The key results from earlier work on outlier detection benchmarking and unsupervised model selection are replicated in our study: model selection remains a critical challenge. A notable discrepancy between our study and earlier benchmarks is the performance of ensembling. A probabilistic ensemble provides the best performance in our study, although its benefits are statistically insignificant over an un-tuned single algorithm, iForest. In domain-agnostic outlier detection benchmarks, ensembling falls short of meta-learning. We remark on several areas of future research.

\begin{table}
\centering

\begin{tabular}{lrrr}
\toprule
trial & MePE (ours) & iForest & Ensemble-P \\
\midrule
compass & 0.645 & 0.649 & \textbf{0.653} \\
hipattack & 0.542 & 0.572 & \textbf{0.577} \\
hope3 & 0.602 & 0.621 & \textbf{0.625} \\
manage & 0.639 & 0.646 & \textbf{0.650} \\
poise & 0.671 & \textbf{0.681} & \textbf{0.681} \\
rely & 0.620 & \textbf{0.629} & 0.618 \\
tips3 & 0.622 & 0.620 & \textbf{0.632} \\
\bottomrule
\label{table:results}
\end{tabular}
\caption{Mean AUROCs of the best meta-learner, single algorithm, and ensemble approach by trial.}
\label{table:results}
\vspace{-4mm}
\end{table}

\noindent \textbf{Preprocessing.} Unlike supervised learning, where preprocessing choices such as imputation scheme can be incorporated as additional hyperparameters and optimized directly for performance, the lack of labels for unsupervised outlier detection renders this impossible. Meta-learning to predict the performance of particular preprocessing techniques is one possible solution \citep{zsautoml}. The preprocessing regime used for this study is simplistic and not tailored to individual algorithms and datasets.

\noindent \textbf{Ensembling.} The success of ensembling warrants further investigation. Earlier benchmarks considered much larger ensembles (300+ models) \citep{zhao2020automating}, and therefore the superior performance in this work may be due to the small size of the ensembles (3-6 models).

\noindent \textbf{Learning regime.} Outlier detection for CSM in MCRCTs is of significant practical interest. Over the course of a trial, irregularity (anomaly) labels become available, opening the door for techniques such as active learning, semi-supervised learning, or other regimes not considered in this study, which only examines the fully unsupervised, offline case. The tradeoffs differ, as well: CSM in MCRCTs is a slow process operating on input data of relatively small dimension not usually exceeding that shown in this study, and does not demand the fast, online, massively scalable solutions required in other domains such as financial ML \citep{zhao2020automating}; rather, maximizing performance is critical.

\acks{This work was supported by the Canadian Institutes for Health Research, Funding Reference Number 175192.}

\bibliography{odrcts}

\appendix

\section{Data Summary}
\label{app:data}

Below, we provide brief information on each of the MCRCTs used in our study. Readers should refer to the original publications for further details:

\begin{itemize}[noitemsep,topsep=0pt,parsep=0pt]
    \item Perioperative Ischemic Evaluation Study (POISE): Randomized 8,351 participants from 190 hospitals in 23 countries undergoing non-cardiac surgery to receive either extended release metoprolol succinate or placebo. \citep{poise}

    \item Randomized Evaluation of Long Term Anticoagulant Therapy (RE-LY): Randomized 18,113 participants from 951 centres in 44 countries with non-valvular atrial fibrillation to receive either of two doses of dabigatran or placebo. \citep{rely} 

    \item Heart Outcomes Prevention Evaluation-3 (HOPE-3): Randomized 12,705 participants from 228 centers in 21 countries to receive candesartan or placebo and rosuvastatin or placebo in a 2-by-2 factorial design. 

    \item The International Polycap Study 3 (TIPS-3): Randomized 5,713 participants from 86 centres in 9 countries without cardiovascular disease but at increased risk of developing it to receive a polypill or placebo, aspirin or placebo, and vitamin D or placebo in a 2-by-2-by-2 factorial design. \citep{tips3}

    \item Rivaroxaban for the Prevention of Major Cardiovascular Events in Coronary or Peripheral Artery Disease (COMPASS): Randomized 27,395 participants from 602 centres in 33 countries with stable atherosclerotic vascular disease to receive either rivaroxaban plus aspirin, rivaroxaban, or aspirin. \citep{compass}

    \item Hip Fracture Accelerated Surgical Treatment and Care Track (HIP ATTACK): Randomized 2,970 participants from 69 hospitals in 17 countries with a hip fracture requiring surgery to receive either accelerated surgery or standard care. 

    \item Management of Myocardial Injury After Noncardiac Surgery (MANAGE): Randomized 1,754 participants from 84 hospitals in 19 countries with myocardial injury following non-cardiac surgery to receive either dabigatran or placebo, and additionally randomized an eligible subset of 556 participants to receive either omeprazole or placebo in a partial 2-by-2 factorial design. \citep{manage}
\end{itemize}

\section{Methods Description}
\label{app:methods}

We briefly describe the meta-learning approaches based on MetaOD \citep{zhao2020automating}, but defer to the original publication for details. MetaOD computes 200 meta-features based on the input dataset alone (no labels required), such as the number of instances or the number of categorical features. The dimensionality of these meta-features is reduced using principal components analysis, and the resulting representations are used as the inputs to a matrix factorization-based recommender system, which learns to rank the algorithms.

MetaOD-P is the pre-trained, publicly release version of MetaOD which has been trained on several open-source anomaly detection benchmark datasets. This version is available on the authors' GitHub: \url{https://github.com/yzhao062/MetaOD}

MetaOD-R is re-trained for each trial in our study using the authors' implementation on our own data in a leave-one-trial-out fashion to eliminate the risk of meta-test set leakage. The script for re-training is available in our code release.

Ensemble-N averages the decision scores of each algorithm as its prediction, while Ensemble-P averages the decision scores scaled to $[0, 1]$ as its prediction, a slight modification that significantly improves performance. Attempts at replicating early comparisons between ensembles and meta-learning suggest that this slight modification has not yet been considered in evaluations of meta-learning approaches for outlier detection.

MePE uses MetaOD-R to predict the ranks of each algorithm, and selects the top-$k$ predicted models, ensembling them as in Ensemble-P.

\section{Evaluation}
\label{app:eval}

\begin{figure}[h]
    \includegraphics[width=8cm]{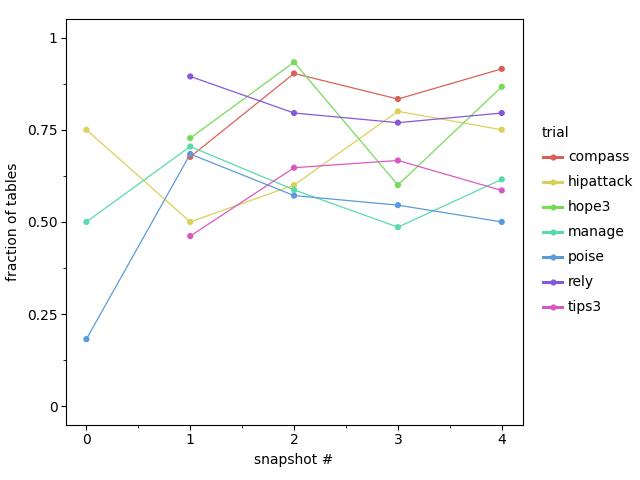}
    \caption{Fraction of datasets from each snapshot that have at least one algorithm achieving positive performance, by trial.}
    \label{fig:success}
\end{figure}

Performance of machine learning-based outlier detection algorithms varies substantially across trials, and less so across snapshots (as the trial progresses), as shown in Figure \ref{fig:success}. In the very early stages of a trial, a very small quantity of data has been collected, accounting for the low success rate in the first snapshot of POISE.

We provide a more granular view of the single best performer, iForest in Table \ref{table:iforest}. As shown, there are no obvious trends that would suggest performance is dependent on the quantity of data collected (the more preliminary the snapshot, the more data is scarce). Additionally, performance across trials is fairly consistent.

\begin{table}
\begin{tabular}{lrrrrr}
\toprule
snapshot \# & 0 & 1 & 2 & 3 & 4 \\
trial &  &  &  &  &  \\
\midrule
compass & - & 0.62 & 0.66 & 0.63 & 0.66 \\
hipattack & 0.66 & 0.56 & 0.57 & 0.59 & 0.54 \\
hope3 & - & 0.66 & 0.63 & 0.57 & 0.62 \\
manage & 0.65 & 0.65 & 0.63 & 0.62 & 0.68 \\
poise & 0.63 & 0.77 & 0.68 & 0.68 & 0.67 \\
rely & - & 0.66 & 0.62 & 0.63 & 0.61 \\
tips3 & - & 0.62 & 0.62 & 0.64 & 0.62 \\
\bottomrule
\end{tabular}
\caption{The performance of iForest by trial and snapshot.}
    \label{table:iforest}
\end{table}

\end{document}